\documentclass[letterpaper]{article}
\usepackage{aaai}
\usepackage{times}
\usepackage{helvet}
\usepackage{courier}
\usepackage{amsfonts, bm}
\usepackage{graphicx, caption, subcaption}

\begin{document}
\title{Domain-Independent Optimistic Initialization for Reinforcement Learning}
\author{Marlos C. Machado,  Sriram Srinivasan and Michael Bowling\\
Reinforcement Learning and Artificial Intelligence Laboratory,\\
Department of Computing Science, University of Alberta\\
\{machado, ssriram, bowling\}@ualberta.ca\\
}
\maketitle
\begin{abstract}
  In Reinforcement Learning (RL), it is common to use optimistic initialization of value functions
  to encourage exploration. However, such an approach generally depends on the domain, viz., 
  the scale of the rewards must be known, and the feature representation must have a constant norm.
  We present a simple approach that performs optimistic initialization with less dependence
  on the domain.
\end{abstract}

\section{Introduction}

One of the challenges in RL is the trade-off between exploration and exploitation. The agent
must choose between taking an action known to give positive reward or to explore other 
possibilities hoping to receive a greater reward in the future. In this context, a common strategy 
in unknown environments is to assume that unseen states are more promising than those states
already seen. One such approach is optimistic initialization of values \cite[Section~2.7]{Sutton98}.

Several RL algorithms rely on estimates of expected values of states or expected values
of actions in a given state \cite{Sutton98}. Optimistic initialization consists in 
initializing such estimates with higher values than are likely to be the true value. To do so,
we depend on prior knowledge of the expected scale of rewards. This paper circumvents
such limitations presenting a different way to optimistically initialize value functions without 
additional domain knowledge or assumptions. 

In the next section we formalize the problem setting as well as the RL framework. We then
present our optimistic initialization approach. Also, we present some experimental analysis of
our method using the Arcade Learning Environment \cite{Bellemare13} as the testbed.
 
\section{Problem Setting}
Consider a Markov Decision Process, at time step $t$ the agent is in a state $s_t \in \mathcal{S}$
and it needs to take an action $a_t \in \mathcal{A}$. Once the action is taken, the agent observes
a new state $s_{t+1}$ and a reward $r_{t+1} \sim R(s_t, a_t, s_{t+1})$ from a transition probability
function $P(s_{t+1} | s_t, a_t) \equiv Pr(s_{t+1} | s_t, a_t)$. The agent's goal is to obtain a policy $\pi(a|s)$ 
that maximizes the expected discounted return $q_\pi(s_t, a_t) \equiv \mathbb{E} 
\Big[\sum_{k=0}^\infty \gamma^k r_{t+k+1} \Big | s_0, \pi \Big]$, where $\gamma \in (0, 1]$ is
the discount factor and $q_\pi(s, a)$ is the action-value function for policy $\pi$. Sometimes it is
not feasible to compute $q_\pi(s, a)$, we then approximate such values with linear function
approximation: $q_\pi(s, a) \approx \theta^T\phi(s,a)$, where $\theta$ is a learned set of weights
and $\phi(s,a)$ is the feature vector. Function approximation adds further difficulties for optimistic
initialization, as one only indirectly specifies the value of state-action pairs through the choice of
$\theta$.

\section{Optimistic Initialization}

An approach to circumvent the requirement of knowing the reward scale is to
normalize all rewards ($r_t$) by the first non-zero reward seen ($r_{1\mbox{\tiny{st}}}$), \emph{i.e.}: 
$r_t/|r_{1\mbox{\tiny{st}}}|$. Then we can optimistically initialize $q_\pi(s,a)$ as $1$, representing the 
expectation that a reward the size of the first reward will be achieved on the next timestep\footnote{
This is only a mild form of optimism. A more optimistic view might be that you can achieve reward on 
each step equal to that of the first observed reward, in which case we should aim to initialize $q_\pi(s,a)$ 
to $\frac{1}{1 - \gamma}$. For sparse reward domains, which is common in the Arcade Learning 
Environment, the mild form is often sufficient.}. With function approximation, this means
initializing the weights $\theta$ to ensure $\theta^T\phi(s_t,a_t) = 1$, \emph{e.g.}: $\theta_i = 1/|\phi(s_t,a_t)|$. 
However, this requires $|\phi(s_t,a_t)|$ to be constant among all states and actions. If the feature vector
is binary-valued then one approach for guaranteeing $\phi$ has a constant norm is to stack $\phi(s_t, a_t)$
and $\neg \phi(s_t, a_t)$, where $\neg$ is applied to each coordinate. While this achieves the goal, it has the
cost of doubling the number of features. Besides, it removes sparsity in the feature vector, which can often be
exploited for more efficient algorithms.

Our approach is to shift the value function so that a zero function is in fact optimistic. We normalize by the
first reward as described above. In addition, we shift the rewards downward by $\gamma - 1$, so 
$\tilde{r}_t = \frac{r_t}{|r_{1\mbox{\tiny{st}}}|} + (\gamma - 1)$. Thus, we have:

\begin{eqnarray*}
  \tilde{q}_\pi(s_t,a_t) &=& \mathbb{E}_\pi\Bigg[\sum_{k=0}^\infty \gamma^k \tilde{r}_{t + k + 1}\Bigg]\\
   &=& \underbrace{\mathbb{E}_\pi\Bigg[\sum_{k = 0}^\infty \gamma^k \frac{r_{t+k+1}}{|r_{1\mbox{\tiny{st}}}|}
   \Bigg]}_{\frac{q_\pi(s_t,a_t)}{|r_{1\mbox{\tiny{st}}}|}} + \underbrace{\sum_{k = 0}^\infty \gamma^k (\gamma - 1)}_{-1}
\end{eqnarray*}

Notice that since $\tilde{q}_\pi(s_t,a_t) = \frac{q_\pi(s_t, a_t)}{|r_{1\mbox{\tiny{st}}}|} - 1$, 
initializing $\theta = 0$ is the same as initializing $q_\pi(s_t, a_t) = r_{1\mbox{\tiny{st}}}$.
This shift alleviates us from knowing $|\phi(s,a)|$, since we do not have the requirement 
$\theta^T \phi(s,a) = 1$ anymore. Also, even though $\tilde{q}_\pi(s_t,a_t)$ is defined in 
terms of $r_{1\mbox{\tiny{st}}}$, we only need to know $r_{1\mbox{\tiny{st}}}$ once a 
non-zero reward is observed. 

In episodic tasks this shift will encourage agents to terminate episodes as fast as possible to 
avoid negative rewards. To avoid this we provide a termination reward $r_{end} =
\gamma^{T-k+1} -1$, where $k$ is the number of steps in the episode and $T$ is the maximum
number of steps. This is equivalent to receiving a reward of $\gamma -1$ for additional $T-k+1$
steps, and forces the agent to look for something better.

\section{Experimental Analysis}
We evaluated our approach in two different domains, with different reward
scales and different number of active features. These domains were obtained from the Arcade Learning 
Environment \cite{Bellemare13}, a framework with dozens of Atari 2600 games where the agent 
has access, at each time step, to the game screen or the RAM data, besides an additional reward
signal. We compare the learning curves of regular Sarsa($\lambda$) \cite{Sutton98} and 
Sarsa($\lambda$) with its Q-values optimistically initialized. We used \emph{Basic} features with
the same Sarsa($\lambda$) parameters reported by \citeauthor{Bellemare13}. The \emph{Basic} 
features divide the screen in to $14 \times 16$ tiles and check, for each tile, if each of the 128
possible colours are active, totalling 28,672 features.

The results are presented in Figure~1. We report results using two different learning rates
$\alpha$, a low value ($\alpha = 0.01$) and a high value ($\alpha = 0.50$), each point corresponds 
to the average after 30 runs.

The game \textsc{Freeway} consists in controlling a chicken that needs to cross a street, avoiding 
cars, to score a point ($+1$ reward). The episode lasts for 8195 steps  and the 
agent's goal is to cross the street as many times as possible. This game poses an interesting 
exploration challenge for ramdom exploration because it requires the agent to cross the street
acting randomly ($|\mathcal{A}|=18$) for dozens of time steps. This means frequently selecting the
action ``go up'' while avoiding cars. Looking at the results in Figure~1 we can see that, as expected,
optimistic initialization does help since it favours exploration, speeding up the process of learning
that a positive reward is available in the game. We see this improvement over Sarsa($\lambda$) for 
both learning rates, with best performance when $\alpha = 0.01$.

The game \textsc{Private Eye} is a very different domain. In this game the agent is supposed 
to move right for several screens (much more than when crossing the street in the game
\textsc{Freeway}) and it should avoid enemies to avoid negative rewards. Along the path the
agent can collect intermediate rewards ($+100$) but its ultimate goal is to get to the end and 
reach the goal, obtaining a much larger reward. We can see that the optimistic initialization is 
much more reckless in the sense that it takes much more time to realize a specific state is not good 
(one of the main drawbacks of this approach), while Sarsa($\lambda$) is more conservative. 
Interestingly, we observe that exploration may have a huge benefit in this game as a larger learning 
rate guides the agent to see rewards in a scale that was not seen by Sarsa($\lambda$).

 \begin{figure}[t]
   \centering
 \begin{subfigure}{0.47\columnwidth}
 \includegraphics[width=\textwidth]{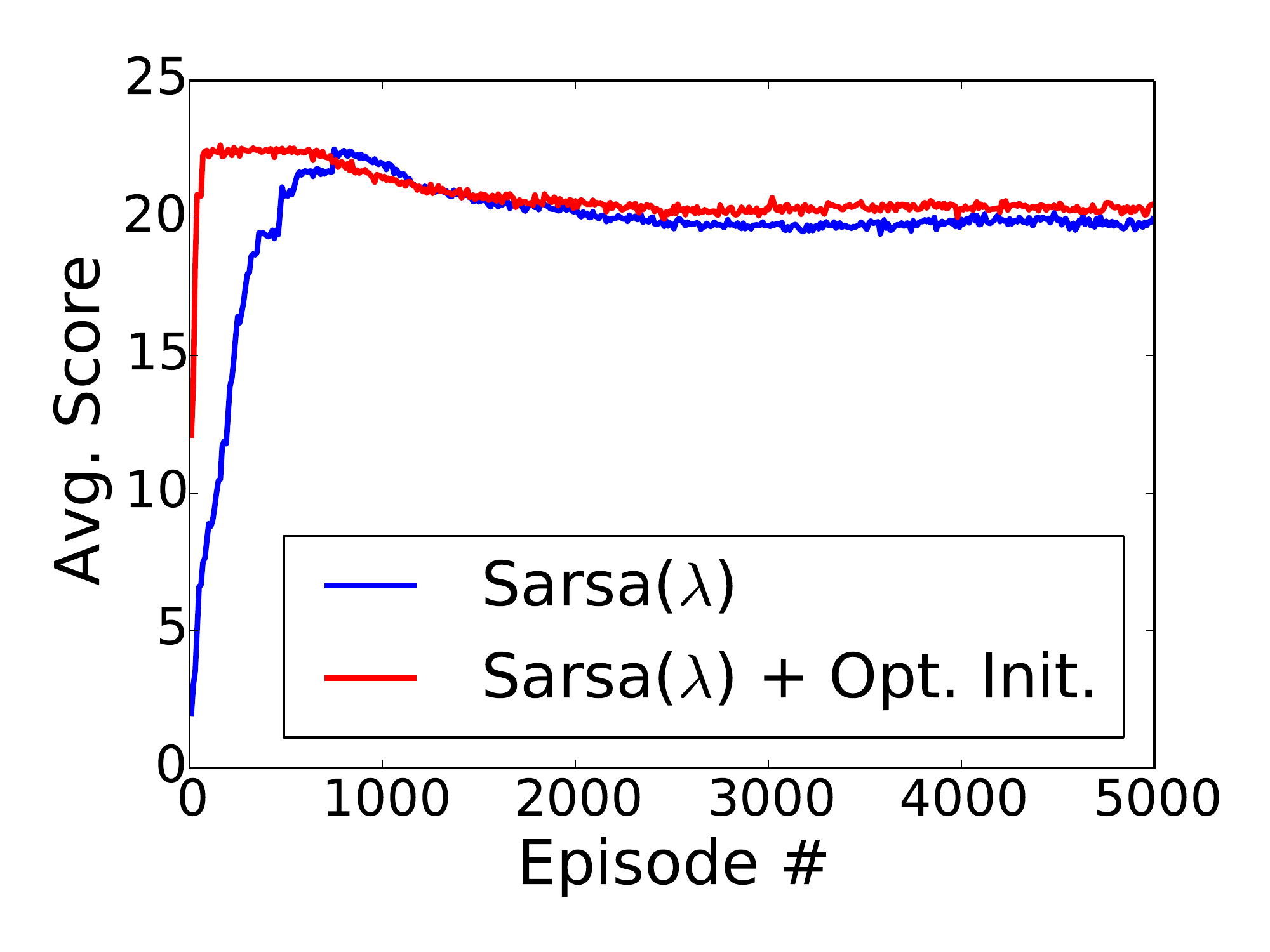}
 \caption{\textsc{Freeway}; $\alpha = 0.01$}
 \end{subfigure}
 ~
  \begin{subfigure}{0.47\columnwidth}
 \includegraphics[width=\textwidth]{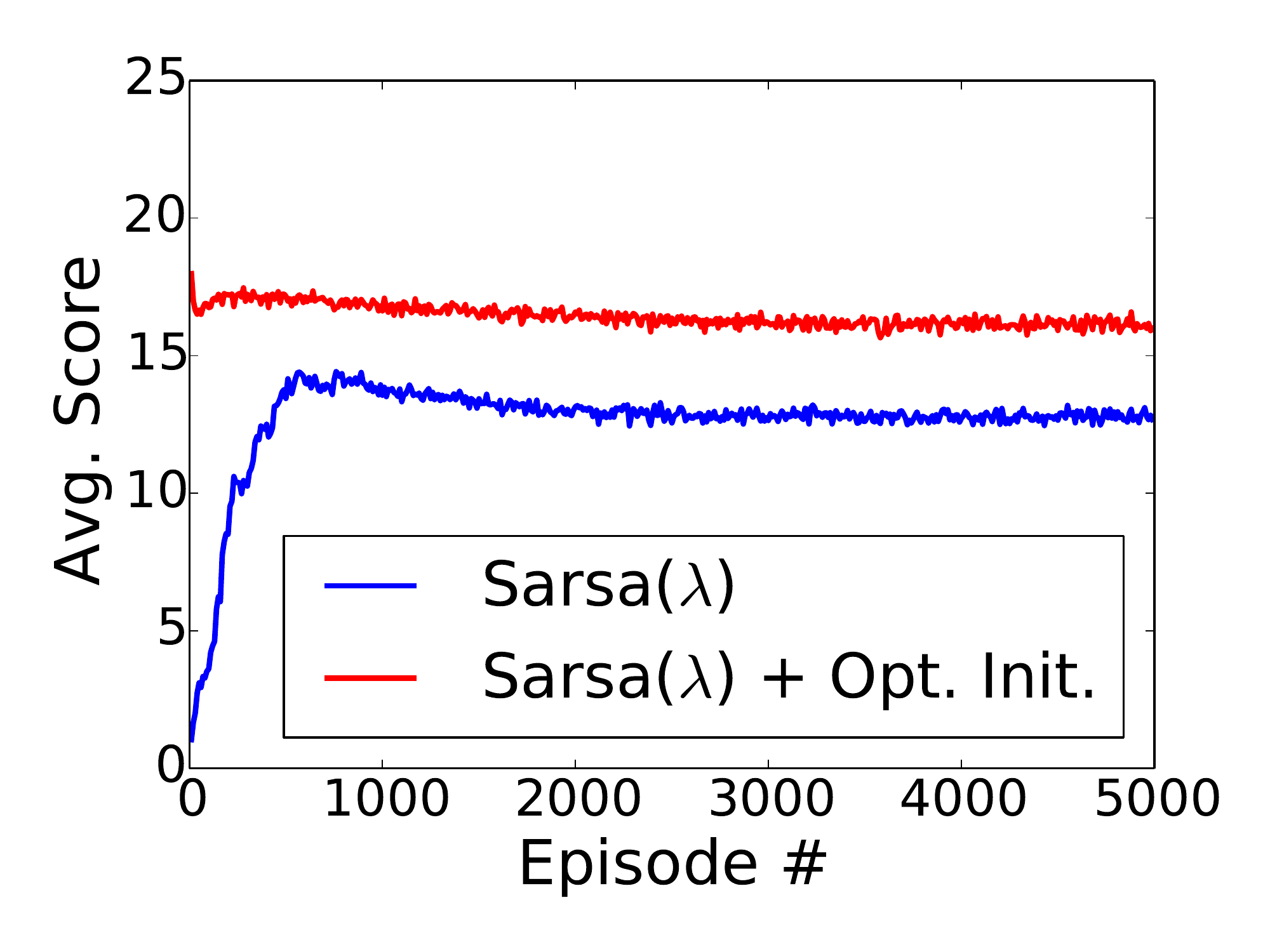}
 \caption{\textsc{Freeway}; $\alpha = 0.50$} 
 \end{subfigure}
 ~
  \begin{subfigure}{0.47\columnwidth}
 \includegraphics[width=\textwidth]{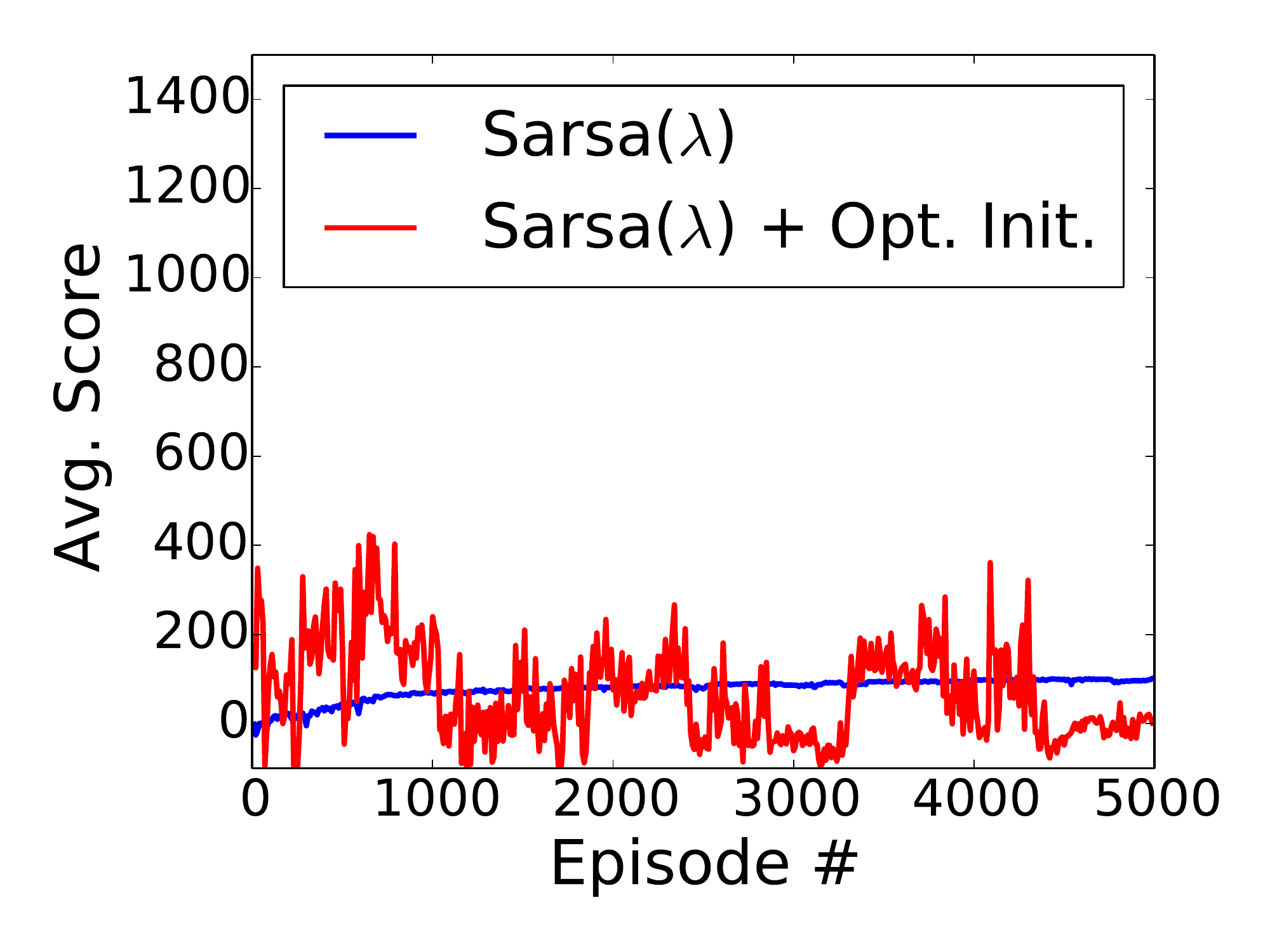}
  \caption{\textsc{Private Eye}; $\alpha = 0.01$}
 \end{subfigure}
 ~
  \begin{subfigure}{0.47\columnwidth}
 \includegraphics[width=\textwidth]{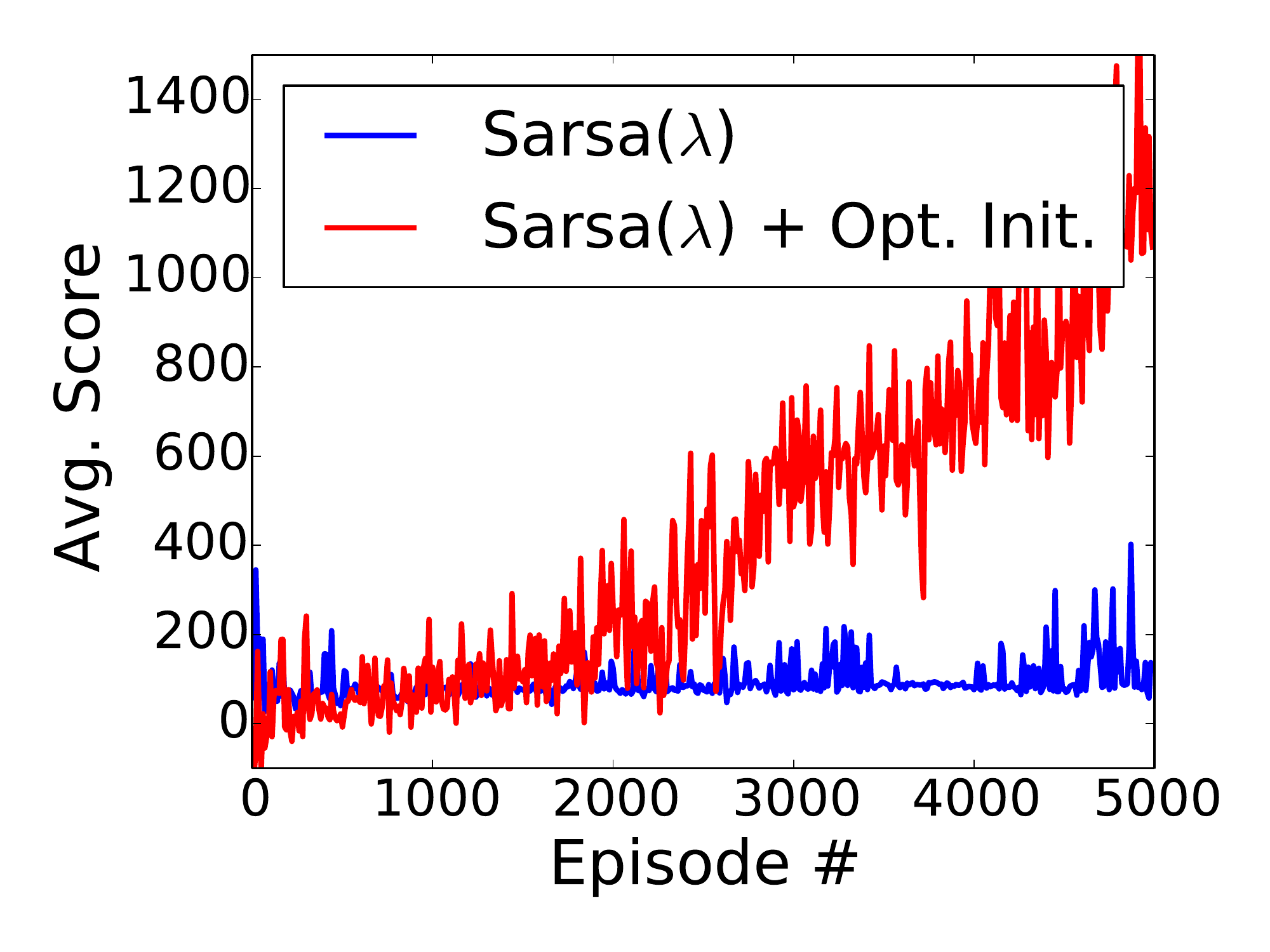}
 \caption{\textsc{Private Eye}; $\alpha = 0.50$} 
 \end{subfigure}
 \caption{Results: Each point corresponds to the average score of the last
 10 episodes \emph{i.e.}, sliding window of size 10.}
\end{figure}

Thus, besides our formal analysis, we have shown here that our approach behaves as one would
expect optimistically initialized algorithms to behave. It increased agents' exploration 
with the trade off that sometimes the agent ``exploited'' a negative reward hoping to obtain a
higher return.

\section{Conclusion}

RL algorithms can be implemented without needing rigorous domain knowledge,
but as far as we know, until this work, it was unfeasible to perform optimistic initialization in
the same transparent way. Besides not requiring adaptations for specific domains, our approach
does not hinder algorithm performance.

\section*{Acknowledgements}
The authors would like to thak Erik Talvitie for his helpful input throughout this research. This 
research was supported by Alberta Innovates Technology Futures and the Alberta Innovates
Centre for Machine Learning and computing resources provided by Compute Canada through Westgrid.

\bibliographystyle{aaai}

\end{document}